\documentclass{article}
\usepackage{spconf,amsmath,graphicx}
\usepackage{flushend}
\usepackage{color}
\usepackage{subfig}
\usepackage{cite}
\usepackage{amsmath}

\def\EPA{\text{XceptionTime}}
\def\minmax{\text{Minmax}}

\title{$\EPA$: A Novel Deep Architecture based on Depthwise Separable Convolutions for Hand Gesture Classification}
\name{Elahe Rahimian$^{\dagger}$, Soheil Zabihi$^\ddagger$, Seyed Farokh Atashzar$^{\dagger\dagger}$, Amir Asif$^\ddagger$, and Arash Mohammadi$^{\dagger}$}

\address{$^\dagger$Concordia Institute for Information System Engineering, Concordia University, Montreal, QC, Canada\\
$^\ddagger$Electrical and Computer Engineering,  Concordia University, Montreal, QC, Canada\\
$^{\dagger\dagger}$Electrical \& Computer Engineering, Mechanical \& Aerospace Engineering, New York University, USA}

\begin{document}
\ninept
\frenchspacing
\maketitle
%
\begin{abstract}
Capitalizing on the need for addressing the existing challenges associated with gesture recognition via sparse multichannel surface Electromyography (sEMG) signals, the paper proposes a novel deep learning model, referred to as the XceptionTime architecture. The proposed innovative XceptionTime is designed by integration of depthwise separable convolutions, adaptive average pooling, and a novel non-linear normalization technique. At the heart of the proposed architecture is several XceptionTime modules concatenated in series fashion designed to capture both temporal and spatial information-bearing contents of the sparse multichannel sEMG signals without the need for data augmentation and/or manual design of feature extraction. In addition, through integration of adaptive average pooling, Conv1D, and the non-linear normalization approach, XceptionTime is less prone to overfitting, more robust to temporal translation of the input, and more importantly is independent from the input window size. Finally, by utilizing the depthwise separable convolutions, the XceptionTime network has far fewer parameters resulting in a less complex network. The performance of XceptionTime is tested on a sub Ninapro dataset, DB1, and the results showed a superior performance in comparison to any existing counterparts. In this regard, $5.71\%$ accuracy improvement, on a window size $200ms$, is reported in this paper, for the first time.
\end{abstract}
%
%
\vspace{-.15in}
\section{Introduction} \label{intro}
\vspace{-.1in}
Recent evolution in deep learning architectures coupled with advancements in rehabilitation technologies has resulted in a promising future to develop intuitive myoelectric prostheses. The surface Electromyography (sEMG) signals~\cite{Dario:SPM, Dario:sEMG, Dario2} derived from the muscle fibers' action potentials, have been used in the literature for hand motion recognition in advanced myoelectric prostheses. In this regard, gesture recognition and classification has attracted a great deal of interest of many researchers due to the high potential for improving the quality of control over the actions of prostheses, which can significantly enhance the quality of lives of hand amputated individuals.

The sEMG signals can be collected based on \textit{sparse multichannel sEMG} or in more advanced cases using \textit{high-density sEMG (HD-sEMG)} devices~\cite{Geng2016, Wei2017, PLOS2018}. Multichannel system records electrical activity of muscles through a spatially distributed electrodes over stump muscles to extract temporal information regarding muscle activity. Multi-channel recording secures several advantages including the ability to obtain large amounts of data from different locations on the muscles which enhances the sparsity of the information space regarding the activities of distributed motor units in the muscles, which potentially allowing for enhancing the quality of classification. Despite the unique advantages of multichannel recording (such as high density systems), the multiplied size of the recorded information space with a high sampling frequency (which can be as high as $3$KHz and is needed for enhancing the fine control) make the processing computationally demanding, which in turn can add latency to the processing pipeline challenging the real-time implementation (which is imperative for the control of prosthetic systems).

It should be also noted that although the performance of deep learning algorithms can motivate the use for multichannel electrode space, applying/training deep models based on signals obtained from sparse multichannel sEMG devices is very challenging as such datasets are typically shallow. The paper aims at addressing this gap by designing a novel deep architecture with reduced computational burden to achieve high accuracy using sparse multichannel sEMG signals. NinaPro~\cite{DB1, Atzori2014} database, which is the most widely accepted benchmark for sparse multichannel sEMG signal processing,  is utilized to design the proposed novel deep~architecture.

\noindent
\textbf{Prior Research}: A common strategy used for hand gesture recognition is to convert the multichannel sEMG recording over fix time windows into images and then use Convolutional Neural Networks (CNN)-based image classification models~\cite{Geng2016, Wei2017, PhyCS, Atzori2016} to perform the recognition task. The problem with such an approach is that only the spatial information of sEMG signals are captured without considering the sequential nature of the sEMG signals. Motivated by this fact, Reference~\cite{PLOS2018} proposed a hybrid CNN and Recurrent Neural Network (RNN) architecture where both spatial and temporal features of the sEMG signals are captured. However, in~\cite{PLOS2018}, raw signals are first converted to images (via six sEMG image representation approaches) and then fed to the hybrid CNN-RNN architecture. The results obtained in~\cite{PLOS2018} show that accuracy in classification depends critically on the characteristic of the constructed images, revealing that there is still a major question what is the optimal approach for converting sEMG signals into images and if this is subject dependent~\cite{Hilbert2019}. Moreover, in this work, the algorithm proposed in Reference \cite{Activity} is utilized, which fuses various signal sequences as an activity image used for training purposes. Although utilization of the aforementioned algorithm allows each sEMG sequence to be adjacent to all other sequences, this requires readjustment of the input signals adding to the complexity of the model. To overcome these problems, we have recently~\cite{Global_elahe} developed a new composite architecture to eliminate the need for converting the raw sEMG signals into images. Instead, these new approaches directly fed the sEMG signals into their proposed temporal-convolutional network architectures capitalizing on the time-series nature of the underlying signals. Although the approach have advantages, i.e., there is no need for readjustment, and the number of parameters is much less than their counterparts using RNN modules, high accuracy can only be achieved by using the complete sEMG sequence (a large window of sEMG sequence). On the other hand, the model in~\cite{ICASSP2019} is trained separately for each subject limiting its generalization capabilities to be used as a subject-independent model. Finally, in~\cite{TBE2019}, the authors extracted $11$ classical sEMG feature sets and then combined these features with a CNN framework. Although this can help with the computational expense of the technique, extraction of optimal engineered features and construction of optimal classifier are particularly challenging and can saturate the achievable accuracy in many cases.

\vspace{.025in}
\noindent
\textbf{Contributions}: The paper aims to address the above-mentioned drawbacks of existing solutions capitalizing on the fact that the problem of recognizing a large set of hand gestures is still far from being solved using sparse multichannel sEMG signals,  both in terms of the recognition accuracy and the complexity of the system. In this regard, we aim to design a novel deep-learning model to classify $52$ hand movements from raw sparse multichannel sEMG signals, without any additional information (such as in~\cite{PLOS2018}), data augmentation (such as in~\cite{ICASSP2019}) or manual design of feature extraction (such as in~\cite{TBE2019}). The paper proposes a novel CNN architecture, which is constructed based on an innovative module, referred to as \textit\textbf{{The XceptionTime}}. The algorithm is designed using the concept of the Inception Networks~\cite{InceptionTime, Xception}. In the proposed architecture, several XceptionTime modules are deployed to classify the hand gesture recognition where both temporal and spatial information-bearing contents of the sparse multichannel sEMG signals are captured. The proposed novel architecture is independent of the window size. This means that by changing the size of the input sequence there is no need to change/reconfigure the architecture itself (in existing deepnet solutions, this is required due to incorporation of fully connected layers within the architecture). To achieve this goal, in the proposed architecture we employed \textit{Adaptive Average Pooling} in the classification layer, which is less prone to over-fitting than traditionally-used fully connected layers~\cite{Networks}. Moreover, a novel method for normalization of the input inspired from~\cite{wavenet} is proposed resulting in better performance both in terms of accuracy and the training speed. Finally, by utilizing the Depthwise Separable  Convolutions, our network has far fewer parameters compared to situation when we use Conv1D convolutions~\cite{Global_elahe}, resulting in less complex network. The proposed algorithm is tested on DB1 sub-database from NinaPro and an accuracy of $93.91\%$ is achieved which is significantly superior to its counterparts in the literature on the same dataset.  

\vspace{-.15in}
\section{Material and Methods} \label{material}
\vspace{-.1in}
In this section, first, the database on which the proposed model is evaluated is described. Then, the pre-processing approach for preparing the data set will be explained.

\vspace{.05in}
\noindent
\textbf{2.1. Database}

\noindent
As stated previously in Section~\ref{intro}, performance of deep learning techniques using sparse multichannel sEMG is yet far being optimal in terms of (i) Recognition accuracy,  (ii) Complexity of the system, and; (iii) Sufficiency of number of subjects and movements. Therefore, the proposed architecture will be evaluated on a public identified scientific benchmark database, Ninapro~\cite{DB1, Atzori2014}, which is the most widely accepted benchmark for evaluation of different models developed based on sparse multichannel sEMG signals. The first Ninapro database~\cite{DB1, Atzori2014}, referred to as the DB1, is used in this work, where the sEMG signals are acquired using Otto Bock MyoBock 13E200 with $10$ wireless electrodes (channels) at a sampling rate of $100$Hz. The DB1 consists of 27 intact (healthy) subjects, where each subject has to repeat $52$ gestures including finger, hand, and wrist movements. The subjects repeated each gestures $10$ times, each time lasted for $5$ seconds followed by $3$ seconds of rest. For the sake of comparison and following the recommendations provided by the database and also previous studies~\cite{Geng2016, Wei2017, Atzori2014, Atzori2016}, the testing set consists of repetitions $2$, $5$, and $7$, where the remaining repetitions are considered as the training set. Evaluating the proposed model based on the sufficient number of subjects and hand gestures, shows its capability to generalize the results for practical use in daily life.

\vspace{.05in}
\noindent
\textbf{2.2. Preprocessing Step}

\setlength{\textfloatsep}{0pt}
\begin{figure}[t!]
\begin{minipage}[b]{.48\linewidth}
\centering
\centerline{\includegraphics[width=4.8cm]{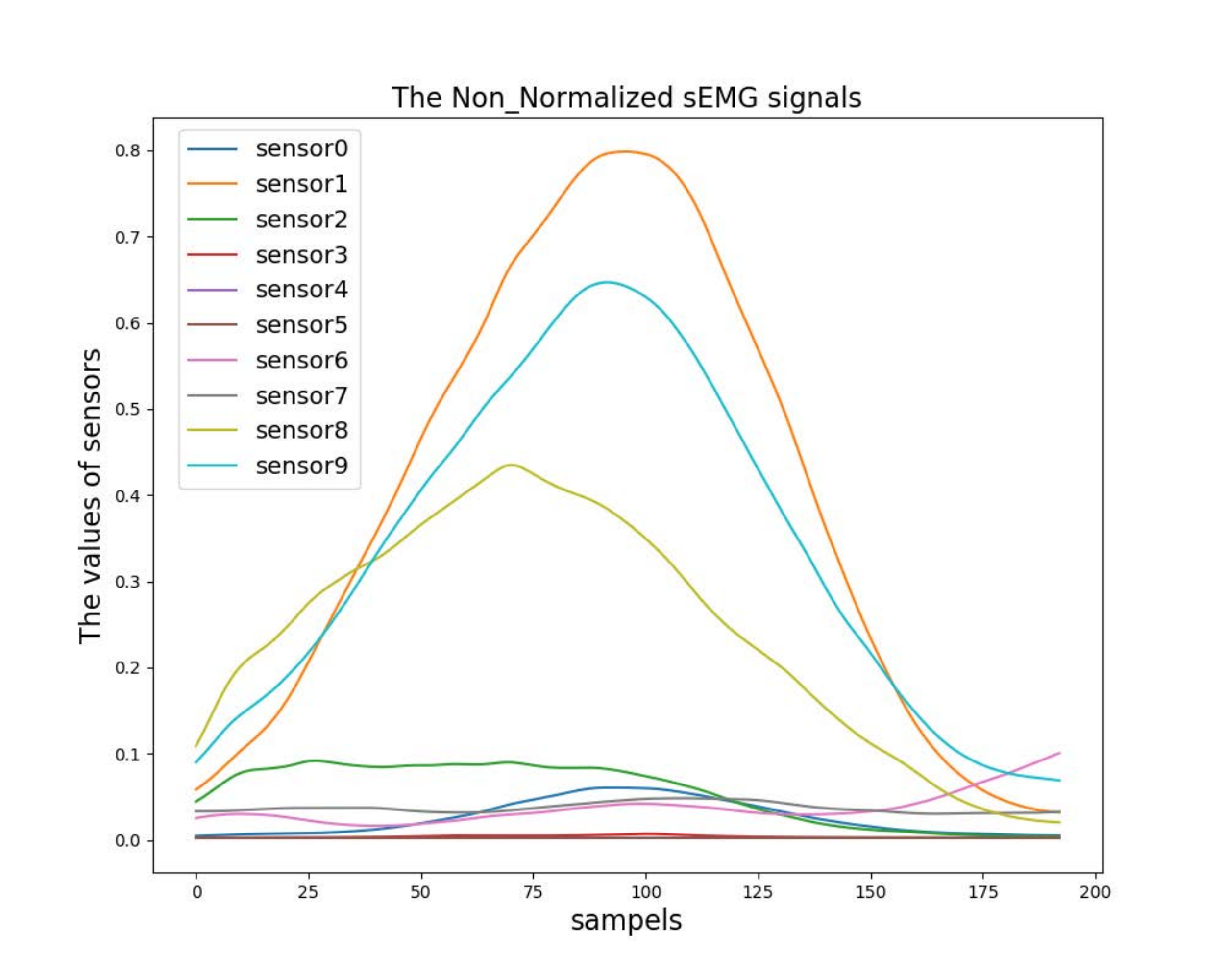}}
\vspace{0cm}
\centerline{(a)}\medskip
\end{minipage}
\hfill
\begin{minipage}[b]{.48\linewidth}
\centering
\centerline{\includegraphics[width=4.8cm]{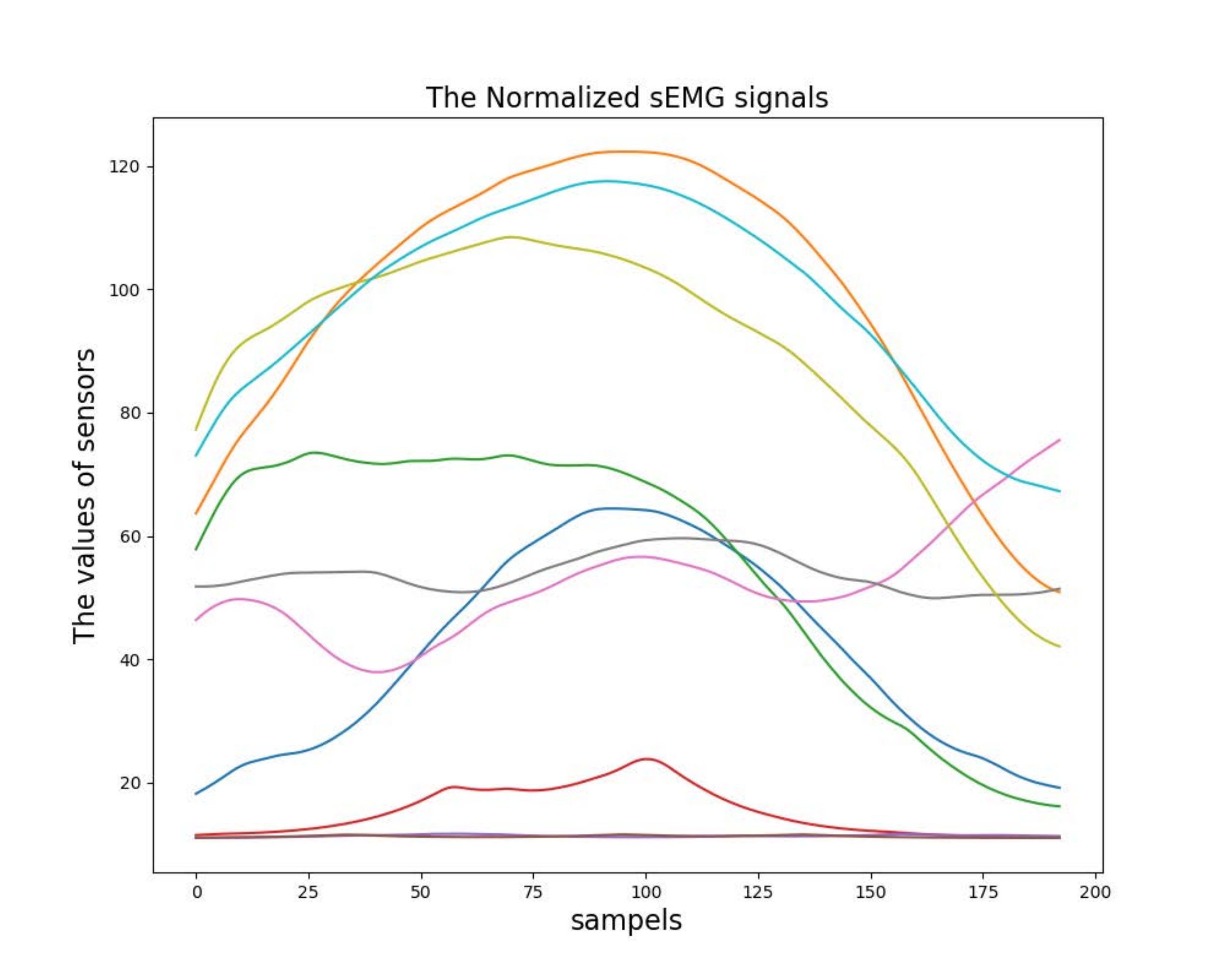}}
\centerline{(b)}\medskip
\end{minipage}
\vspace{-.5cm}
\caption{\footnotesize The electrical activity of muscles obtained from $10$ sensors: (a) The sEMG signals before normalization. (b) The sEMG signals after $\mu$-law nonlinear normalization.}
\label{norm}
\end{figure}
\begin{figure*}[htb]
\begin{minipage}[b]{1.0\linewidth}
\centering
\centerline{\includegraphics[width=14cm]{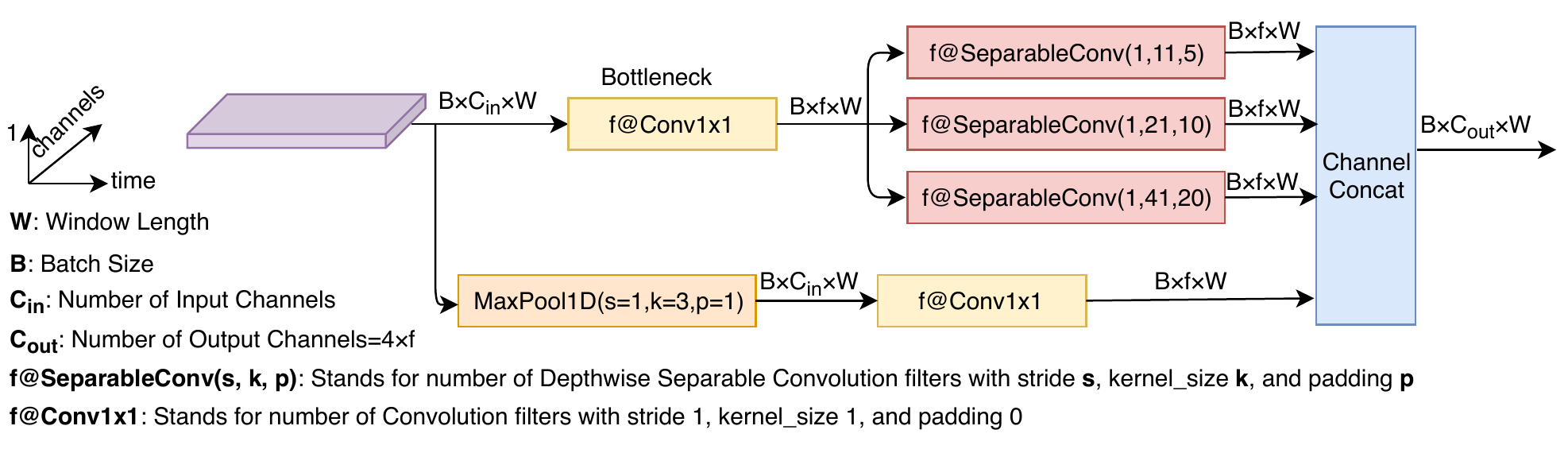}}
\centerline{(a) XceptionTime Module}\medskip
\end{minipage}
\begin{minipage}[b]{1.0\linewidth}
\centering
\centerline{\includegraphics[width=14cm]{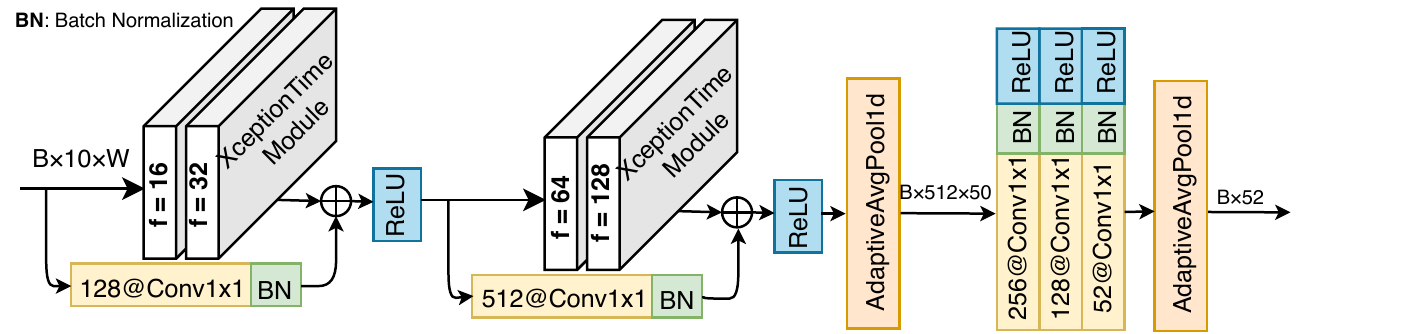}}
\centerline{(b) XceptionTime Architecture}\medskip
\vspace{-.15in}
\end{minipage}
\caption{\footnotesize (a) \textit{\textbf{XceptionTime Module}}, which consists of two parallel paths, the first path includes three Depthwise Separable Convolutions, while the second path includes a MaxPooling followed by a Conv$1\times1$. (b) \textit{\textbf{$\EPA$ Architecture}}, which includes series of XceptionTime modules with residual connections followed by Adaptive Average Pooling layers and Conv$1\times1$ layers.}
\label{archit}
\vspace{-0.3in}
\end{figure*}
\noindent
Following the proposed preprocessing procedure in the previous studies~\cite{Geng2016, Wei2017,Atzori2014,Atzori2016}, we adopted a $1^{\text{st}}$ order $1$Hz low-pass Butterworth filter to preprocess the electrical activities of muscles. However, we develop and propose a new approach for the normalization, referred to as the $\mu$-law normalization, of sEMG signals in a nonlinear fashion based on $\mu$-law transformation \cite{wavenetITUT}. This normalization approach has been used traditionally in speech and communication domains for quantization purposes. We propose for the first time to use it for normalization in the context of sEMG processing. The $\mu$-law normalization is performed based on the following formulation:
\begin{equation}\label{mu_law}
F(x_t) = \text{sign}(x_t)\frac{\ln{\big(1+ \mu |x_t|\big)}}{\ln{\big(1+ \mu \big)}}.
\vspace{-.1in}
\end{equation}
where $x_t$ denotes the input scaler to be normalized, and $\mu = 256$ is utilized. The nonlinear normalization preprocesses the sEMG signals significantly better than linear normalization such as $\minmax$ normalization, which is commonly used. In contrary to commonly used $\minmax$ normalization, which linearly distributed signal values between the pre-defined range, the proposed $\mu$-law normalization magnifies the outputs of sensors with small magnitude (in a logarithmic fashion), while keeping the scale with those sensors having larger values over time.
As an illustrative example, Fig.~\ref{norm} shows the 1Hz low-pass filtered sEMG signals obtained from $10$ sensors corresponding to the first  repetition from Subject $1$ performing the second gesture. As can be observed in Fig.~\ref{norm}(a), except from sensors  $1$, $8$, and $9$, the values of the remaining sensors are close to zero. However, by using the proposed $\mu$-law normalization (as shown in Fig.~\ref{norm}(b)), the outputs of the sensors will be amplified more nonlinearly.

\vspace{-.15in}
\section{The Proposed $\EPA$ Architecture} \label{architecture}
\vspace{-.1in}

In~\cite{InceptionTime}, inspired by the Inception V4 architecture, a new deepnet model has been recently proposed and named as ``InceptionTime''  for time series classification. In~\cite{InceptionTime} it is shown that InceptionTime, which is an equivalent of AlexNet for time series data, is more accurate and faster than its existing counterparts in time series classification. On the other hand, in~\cite{Xception}, by replacing the Inception modules with depthwise separable convolution, a new architecture is designed and named as Xception, which has  better performance than Inception V3 on a large image classification dataset. Motivated by the prior works~\cite{InceptionTime, Xception}, we propose a novel deep architecture, the $\EPA$, which is more accurate than the existing model for sparse sEMG-based hand gesture recognition. Furthermore, by deploying adaptive average pooling, the proposed end-to-end $\EPA$ architecture is independent of the time window, meaning that for utilization of different time windows, e.g., $50$ms, $100$ms, or $150$ms, there is no need to reconfigure and retrain the $\EPA$ model. Besides, by replacing the fully connected layers with adaptive average pooling, the proposed $\EPA$ model is less prone to overfitting because there are no extra parameters to optimize~\cite{Networks}. By deploying adaptive average pooling, the proposed  architecture is more robust to temporal translation of the inputs as the temporal information will sum out.

In the following sub-sections, first, the proposed $\EPA$ module is introduced followed by a description of the $\EPA$ architecture consisting of stacked $\EPA$ modules and adaptive average pooling layers.

\vspace{.05in}
\noindent
\textbf{3.1. XceptionTime Module}\label{module}

\noindent
One of the challenging tasks in designing CNNs, is selecting the right kernel size, which has an important role in extracting global or local information. However, inspired by Inception~\cite{inception}, as shown in Fig.~\ref{archit}(a), instead of committing ourselves to pick a filter with a specific size, we adopt multiple one-dimensional filters with different kernel sizes to extract short and long time series' features simultaneously with the resulted feature maps being concatenated to construct the output features. Moreover, for mitigating the computational cost problems, as well as lessening the overfitting problems, the \textit{bottleneck} layer is used as the first component within the proposed $\EPA$ Module. In the bottleneck layer, \textit{f} number of one-dimensional filters with kernel size one is utilized to transform the input with $C_{in}$ channels into another time series with \textit{f} channels.

One key difference between the proposed $\EPA$ module and InceptionTime module previously proposed in~\cite{InceptionTime}, is deploying depthwise separable convolutions, which significantly mitigates the required number of parameters in the network.  In Depthwise Separable Convolution~\cite{DSC, DSC2}, two convolutions are deployed, i.e., the \textit{Depthwise Convolution}, and the \textit{Pointwise Convolution}. In Depthwise Convolution, each channel of the input is convolved separately and then stacked together; therefore, the temporal convolution is done without changing the depth. The consequent output from the Depthwise convolution is fed to the Pointwise convolution, where $1\times 1$ convolutions are utilized to transform the number input channels from the Depthwise convolution into a new channel depth. Later in Section \ref{results}, it will be shown that by using the depthwise separable convolutions, not only the recognition accuracy will be increased, but also the number of parameters will be reduced significantly.

To summarize, as shown in Fig.~\ref{archit}(a), the time series input with $C_{in}$ number of channels is first fed to two parallel paths. The first path consists of a bottleneck, reducing the dimensionality of the input, followed by three sets of depthwise separable convolution each with \textit{f} number of filters with kernel size \textit{l}, where \textit{l} is set to $11$, $21$, or $41$. In the second path, the input is fed to a MaxPooling layer followed by a Conv$1\times1$ component, which produces an output with \textit{f} channels. Finally, the resulted feature maps of Depthwise Separable Convolutions and skip connections are concatenated in a channel-wise fashion.  As shown in Fig.~\ref{archit}(a), the time series input with C$_{in}$ channels are transformed to output with $C_{out}$ number of channels, where $C_{out}$ is four times that of the number of filters ($f$) used in the bottleneck as well as in the depthwise separable convolutions.

\vspace{.05in}
\noindent
\textbf{3.2. XceptionTime Architecture}

\noindent
The  $\EPA$ architecture is constructed based on the proposed $\EPA$ modules described in Sub-section 3.1. More specifically, after preprocessing, sEMG signals acquired from $10$ sensors are segmented by a window with a length of $W$ $\in \{50ms, 100ms, 150ms, 200ms\}$ (it is worth mentioning that $W$ should be under $300$ms to satisfy the acceptable delay time~\cite{300}). The sliding window with steps of $10$ms is considered for segmentation of the sEMG signals. The proposed $\EPA$ architecture (Fig.~\ref{archit}(b)), includes $4$ XceptionTime modules where the number of filters ($f$) are set to 16, $32$, $64$, and $128$, respectively. Moreover, two residual connections~\cite{Resnet} are deployed in the $\EPA$ architecture  to address the degradation problem. Each residual connection consists of a Conv$1\times1$ layer, to match up the input and output dimensions, followed by Batch Normalization, which is for regularization and also to reduce the internal covariate shift effect~\cite{Batchnorm}. Moreover, in order to learn the complex structure of data, Rectified Linear Unit (ReLU) is applied to the summation of outputs of residual connection and the $\EPA$ module.

As stated previously, one of the novelties of the proposed $\EPA$ architecture is the independency of the architecture from the length of the time window. In other words, for an arbitrary window length, the $\EPA$ architecture remains unchanged without any need of reconfiguration. To realize this objective, the output yielded from summation of the $4^{\text{th}}$ XceptionTime Module and residual Connection is fed to an Adaptive Average Pooling layer, which transforms the input with window size $W$ to a fixed length of $50$. Then, the dimension of the time series input will be reduced to the number of the classes (i.e., $52$ in our settings) by stacking three Conv$1\times1$, each followed by Batch Normalization and ReLU. Finally, a second Adaptive Average Pooling layer is used to convert the length of the input signal to one. We use Adam optimizer for training purpose with learning rate of 0.001. The learning rate changes in a cycle with a length of 20 epochs. After 20 epoch, we divided the learning rate by $2$. These models are trained with a mini-batch size of $32$. For measuring the classification performance the Cross-entropy loss is considered.

\vspace{-.15in}
\section{Experiments and results} \label{results}
\vspace{-.1in}

In this section, the performance of the proposed architecture is evaluated through a comprehensive set of different  experiments and provide comparisons with $6$ state-of-the-art models~\cite{Geng2016, Wei2017, PLOS2018, Atzori2016, ICASSP2019, TBE2019} developed recently on the same dataset to illustrates superior performance of the proposed $\EPA$ over its counterparts.

\vspace{.025in}
\noindent
\textbf{Experiment 1}: In this experiment, referred to as ``First Exp.'' in the results, the objective is to validate our claim that by incorporation of Depthwise Separable Convolutions within the proposed $\EPA$ architecture, a much smaller model size with significantly reduced complexity will be achieved. For this purpose, we implemented a variant of the proposed architecture, where referred to as $\EPA$-V2, where standard convolutions are deployed within the $\EPA$ Module instead of  Depthwise Separable Convolutions. Table~\ref{table1} shows the results, where it can be observed that while the accuracy associated with the XceptionTime is slightly better than $\EPA$-V2, the number of parameters is significantly reduced. For example, the accuracy for the proposed XceptionTime model for window length $200$ms is $95.43$\% using $ 413,516$ number of parameters, while $\EPA$-V2 archives accuracy of $94.59$\% but using extensively higher $1,918,476$ of parameters.
\begin{table}[t!]
\vspace{-0.1in}
\centering
\caption{\footnotesize (First Exp.): Comparison between the proposed $\EPA$ model and $\EPA$-V2 model. (Second Exp.): Result of the proposed model and $\EPA$-V2 when the input is normalized by $\minmax$.\label{table1}}
\vspace{-0.1in}
\includegraphics[scale=.58]{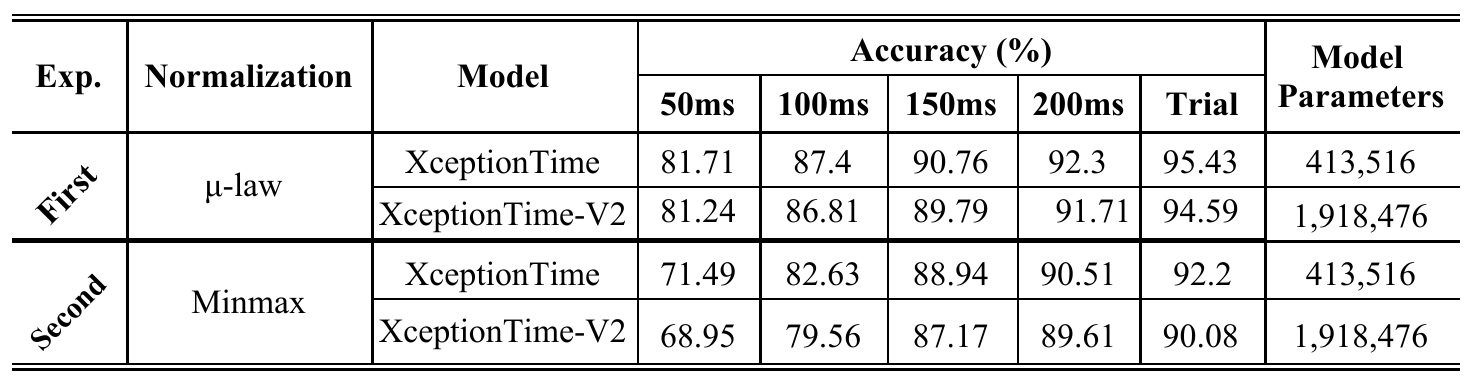}
\vspace{-0.1in}
\end{table}

\vspace{.025in}
\noindent
\textbf{Experiment 2}: In this experiment, referred to as ``Second Exp.'' in the results, the objective is to validate the effectiveness of using the proposed non-linear $\mu$-normalization within the proposed $\EPA$ architecture. In this regard, in Table~\ref{table1}, results trained by using $\minmax$ normalization is shown for both variants of the proposed framework. From  Table~\ref{table1}, it is observed that the accuracy of the model will decrease when $\minmax$ normalization is applied to the input. For instance, accuracy of the proposed XceptionTime framework with $\mu$-law normalization in window length of $50$ms is $81.71$\%, whereas using $\minmax$ normalization within the proposed XceptionTime framework reduces the accuracy to $71.49$\%.  Another observation is that the degradation effect of discarding the proposed nonlinear normalization approach on  $\EPA$-V2 is higher.

\begin{table}[t!]
\vspace{-0.1in}
\centering
\caption{\footnotesize Accuracy when the proposed $\EPA$ Model is trained on a combination of different window lengths (i.e. 50, 100, 150, 200) and then tested on different windows.\label{table2}}
\vspace{-0.15in}
\includegraphics[scale=.8]{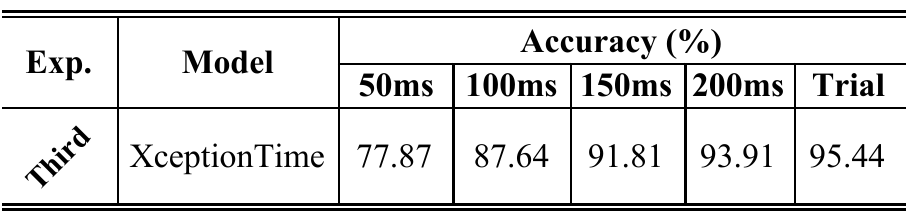}
\end{table}
\vspace{.025in}
\noindent
\textbf{Experiment 3}: The third experiment is performed to validate our claim that the proposed $\EPA$ is applicable to different window sizes without the need for reconfiguration.~We evaluate the performance when the proposed architecture is trained based on a combination of different window sizes. In other words, instead of training the XceptionTime model just with a specific time window (as is done for reporting the results in Table~\ref{table1}), inputs with different window sizes are fed into network to increase the robustness of the network during training. However, for the effectiveness of the training process, only windows with the length of $50$, $100$, $150$, and $200$ are used as input. Table~\ref{table2} illustrates the results obtained from  XceptionTime trained with a combination of different window lengths and then tested separately on each window size. As can be seen, the performance of the model, except for time window $50$, is improved in comparison to the case where the model was just trained with a specific window length (Table\ref{table1}(First Exp.)). In other words, not only the proposed model can handle different window sizes simultaneously, by utilizing this property the performance can be boosted. Finally, Table~\ref{table3} shows performance of the proposed  model in comparison to the state-of-the art results obtained over the same DB1 dataset of $52$ hand gestures. As shown in Table\ref{table3}, our architecture outperforms existing solutions while maintaining a reduced number of parameters.

\begin{table}[t!]
\vspace{-0.1in}
\centering
\caption{\footnotesize Comparison of the proposed $\EPA$l with the state-of-the-art literature (number of parameters for~\cite{Geng2016,Wei2017,Atzori2016} are reported from~\cite{ICASSP2019})\label{table3}.}
\vspace{-0.14in}
\includegraphics[scale=.68]{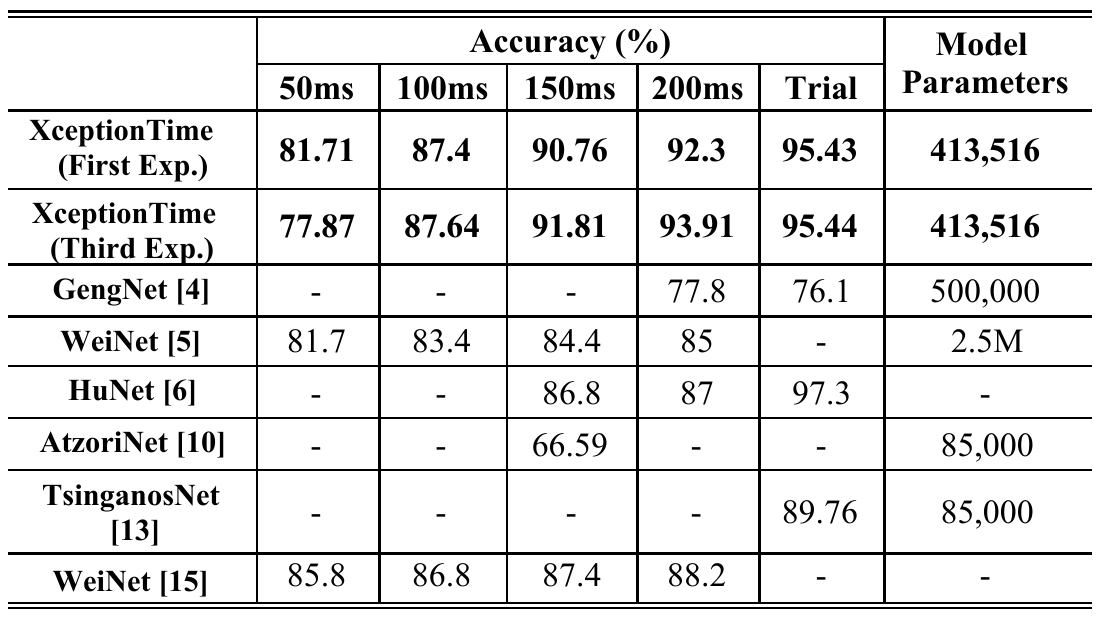}
\end{table}

\vspace{-.15in}
\section{Conclusion}\label{sec:page}
\vspace{-.1in}
With the goal of addressing identified shortcomings of existing models for recognition tasks via sparse multichannel  surface Electromyography (sEMG) signals, the paper proposed the novel  $\EPA$ architecture. The proposed innovative $\EPA$  is designed by integration of depthwise separable convolutions, adaptive average pooling, and a novel non-linear normalization technique. To the best of our knowledge, it is the first time that the proposed innovative $\EPA$ architecture is introduced and has not been designed/utilized previously in any application. Its performance is evaluated via the benchmark sparse sEMG dataset outperforming any existing counterparts. As an attempt to achieve reproducibility, the code will be released on GitHub.

\vfill\pagebreak



\begin{thebibliography}{10}
\bibitem{Dario:SPM}
N. Jiang, S. Dosen, K.R. Muller, D. Farina,
\newblock ``Myoelectric Control of Artificial Limbs-Is There A Need to Change Focus? [in the spotlight],''
\newblock {\em IEEE Signal Process. Mag.}, vol. 29, pp. 150-152, 2012.

\bibitem{Dario:sEMG}
D. Farina, R. Merletti, R.M. Enoka,
\newblock ``The Extraction of Neural Strategies from the Surface EMG,''
\newblock {\em J. Appl. Physiol.}, vol. 96, pp. 1486-95, 2004.

\bibitem{Dario2}
M. Zia ur Rehman, S. Gilani, A. Waris, I. Niazi, G. Slabaugh, D. Farina, E. Kamavuako,
\newblock ``Stacked Sparse Autoencoders for EMG-based Classification of Hand Motions: A Comparative Multi Day Analyses between Surface and Intramuscular EMG,''
\newblock {\em Appl. Sci.}, vol. 8, 1126,  2018.

\bibitem{Geng2016}
W. Geng, Y. Du, W. Jin, W. Wei, Y. Hu, and J. Li,
\newblock ``Gesture recognition by instantaneous surface EMG images. Scientific reports.,''
\newblock {\em Scientific reports 6}, p.36571, 2016.

\bibitem{Wei2017}
W. Wei, Y.Wong, Y. Du, Y. Hu, M. Kankanhalli, and  W. Geng,
\newblock ``A multi-stream convolutional neural network for sEMG-based gesture recognition in muscle-computer interface.,''
\newblock {\em Pattern Recognition Letters}, 2017.

\bibitem{PLOS2018}
Y. Hu, Y. Wong, W. Wei, Y. Du, M. Kankanhalli, and W. Geng,
\newblock ``A Novel Attention-based Hybrid CNN-RNN Architecture for sEMG-based Gesture Recognition,''
\newblock {\em PloS One}, vol. 13, no. 10, 2018.

\bibitem{DB1}
M. Atzori, A. Gijsberts, I. Kuzborskij, S. Heynen, A.G.M Hager, O. Deriaz, C. Castellini, H. Müller, and B. Caputo,
\newblock ``A Benchmark Database for Myoelectric Movement Classification,''
\newblock {\em Transactions on Neural Systems and Rehabilitation Engineering}, 2013.

\bibitem{Atzori2014}
M. Atzori, A. Gijsberts, C. Castellini, B. Caputo, A.G.M. Hager, S. Elsig, G. Giatsidis, F. Bassetto, and H. Müller,
\newblock ``Electromyography data for non-invasive naturally-controlled robotic hand prostheses.,''
\newblock {\em Scientific data 1}, 140053, 2014.

\bibitem{PhyCS}
P. Tsinganos, B. Cornelis, J. Cornelis, B. Jansen, and A. Skodras,
\newblock ``Deep Learning in EMG-based Gesture Recognition.,''
\newblock {\em PhyCS}, pp. 107-114, 2018.

\bibitem{Atzori2016}
M. Atzori, M. Cognolato, and H. Müller,
\newblock ``Deep learning with convolutional neural networks applied to electromyography data: A resource for the classification of movements for prosthetic hands,''
\newblock {\em Frontiers in neurorobotics 10}, p.9, 2016.

\bibitem{Hilbert2019}
P. Tsinganos, B. Cornelis, J. Cornelis, B. Jansen, A. and Skodras,
\newblock ``A Hilbert Curve Based Representation of sEMG Signals for Gesture Recognition,''
\newblock {\em International Conference on Systems, Signals and Image Processing (IWSSIP)}, 201-206, 2019.

\bibitem{Activity}
W. Jiang, and Z. Yin,
\newblock ``Human Activity Recognition using Wearable Sensors by Deep Convolutional Neural Networks,''
\newblock {\em ACM International Conference on Multimedia}, 2015, pp. 1307-1310.

\bibitem{ICASSP2019}
P. Tsinganos, B. Cornelis, J. Cornelis, B, Jansen, and A. Skodras,
\newblock ``Improved Gesture Recognition Based on sEMG Signals and TCN,''
\newblock {\em IEEE International Conference on Acoustics, Speech and Signal Processing (ICASSP)}, 2019, pp. 1169-1173.

\bibitem{Global_elahe}
E. Rahimian, S. Zabihi, S. F. Atashzar, A. Asif, A. Mohammadi,
\newblock ``sEMG-Based Hand Gesture Recognition via Dilated Convolutional Neural Networks,''
\newblock {\em GlobalSIP}, 2019.

\bibitem{TBE2019}
W. Wei, Q. Dai, Y. Wong, Y. Hu, M. Kankanhalli, and W. Geng,
\newblock ``Surface Electromyography-based Gesture Recognition by Multi-view Deep Learning,''
\newblock {\em IEEE Transactions on Biomedical Engineering.}, 2019.

\bibitem{InceptionTime}
H. I. Fawaz, B. Lucas, G. Forestier, C. Pelletier, D. F. Schmidt, J. Weber, G. I. Webb, L. Idoumghar, P. Muller, F. Petitjean,
\newblock ``InceptionTime: Finding AlexNet for Time Series Classification.,''
\newblock {\em arXiv:1909.04939}, 2019.

\bibitem{Xception}
F. Chollet,
\newblock ``Xception: Deep learning with depthwise separable convolutions.,''
\newblock {\em In Proceedings of the IEEE conference on computer vision and pattern recognition }, 2017, pp.1251-1258.


\bibitem{Networks}
M. Lin, Q. Chen, and S. Yan,
\newblock ``Network in network,''
\newblock {\em arXiv preprint arXiv:1312.4400}, 2013.

\bibitem{wavenet}
A.V.D. Oord, S. Dieleman, H. Zen, K. Simonyan, O. Vinyals, A. Graves, N. Kalchbrenner, A. Senior, and K. Kavukcuoglu,
\newblock ``Wavenet: A generative model for raw audio,''
\newblock {\em arXiv preprint arXiv:1609.03499}, 2016.

\bibitem{wavenetITUT}
TU-T. Recommendation G. 711
\newblock ``Pulse code modulation (PCM) of voice frequencies,''
\newblock {\em ITU}, 1988.

\bibitem{inception}
C. Szegedy, W. Liu, Y. Jia, P. Sermanet, S. Reed, D. Anguelov, D. Erhan, V. Vanhoucke, and A. Rabinovich,
\newblock ``Going deeper with convolutions,''
\newblock {\em In Proceedings of the IEEE conference on computer vision and pattern recognition}, 2015, pp. 1-9.



\bibitem{DSC}
L. Sifre, S. and Mallat, S.,
\newblock ``Rigid-motion scattering for image classification,''
\newblock {\em Ph. D. dissertation}, 2014.

\bibitem{DSC2}
V. Vanhoucke,
\newblock ``Learning visual representations at scale,''
\newblock {\em ICLR invited talk}, 2014.

\bibitem{300}
B. Hudgins, P. Parker, and R.N. Scott,
\newblock ``A new strategy for multifunction myoelectric control,''
\newblock {\em IEEE Transactions on Biomedical Engineering} , vol. 40, no. 1, p.82-94, 1993.

\bibitem{Resnet}
K. He, X. Zhang, S. Ren, and J. Sun,
\newblock ``Deep residual learning for image recognition,''
\newblock {\em In Proceedings of the IEEE conference on computer vision and pattern recognition} , 2016, pp. 770-778.

\bibitem{Batchnorm}
S. Ioffe, and C. Szegedy,
\newblock ``Batch normalization: Accelerating deep network training by reducing internal covariate shift,''
\newblock {\em arXiv preprint arXiv:1502.03167} , 2015.


\end{thebibliography}
\end{document}